\documentclass[conference]{IEEEtran}
\IEEEoverridecommandlockouts
\usepackage{cite}
\usepackage{amsmath,amssymb,amsfonts}
\usepackage[ruled,vlined,]{algorithm2e}
\usepackage{algpseudocode}
\usepackage{graphicx}
\usepackage{textcomp}
\usepackage{xcolor}

\usepackage{eso-pic}
\usepackage{lipsum}

\def\BibTeX{{\rm B\kern-.05em{\sc i\kern-.025em b}\kern-.08em
T\kern-.1667em\lower.7ex\hbox{E}\kern-.125emX}}
\newcommand{\linebreakand}{%
\end{@IEEEauthorhalign}
\hfill\mbox{}\par
\mbox{}\hfill\begin{@IEEEauthorhalign}
}

\AddToShipoutPictureBG*{%
  \AtPageUpperLeft{%
    \hspace{\paperwidth}%
    \raisebox{-\baselineskip}{%
      \makebox[0pt][r]{\tiny{\textcopyright 2022 IEEE. Personal use of this material is permitted. Permission from IEEE must be obtained for all other uses, in any current or future media, including reprinting/republishing this material for advertising or promotional purposes, creating new collective works, for resale or redistribution }}
}}}%
\AddToShipoutPictureBG*{%
  \AtPageUpperLeft{%
    \hspace{\paperwidth}%
    \raisebox{-1.5\baselineskip}{%
      \makebox[0pt][r]{\tiny{to servers or lists, or reuse of any copyrighted component of this work in other works.} }
}}}%

\begin{document}

\title{DogTouch: CNN-based Recognition of Surface Textures by Quadruped Robot with High Density Tactile Sensors\thanks{The reported study was funded by RFBR and CNRS, project number 21-58-15006.}
}

\author{

\IEEEauthorblockN{Nipun Dhananjaya \\ Weerakkodi Mudalige}
\IEEEauthorblockA{\textit{ISR Laboratory} \\
\textit{Skoltech}\\
Moscow, Russia \\
nipun.weerakkodi@skoltech.ru}
\and
 
\IEEEauthorblockN{Elena Nazarova}
\IEEEauthorblockA{\textit{ISR Laboratory} \\
\textit{Skoltech}\\
Moscow, Russia \\
elena.nazarova@skoltech.ru}
\and
 
\IEEEauthorblockN{Ildar Babataev}
\IEEEauthorblockA{\textit{ISR Laboratory} \\
\textit{Skoltech}\\
Moscow, Russia \\
ildar.babataev@skoltech.ru}
\and
 
\IEEEauthorblockN{Pavel Kopanev}
\IEEEauthorblockA{\textit{ISR Laboratory} \\
\textit{Skoltech}\\
Moscow, Russia \\
pavel.kopanev@skoltech.ru}
\and

\IEEEauthorblockN{Aleksey Fedoseev}
\IEEEauthorblockA{\textit{ISR Laboratory} \\
\textit{Skoltech}\\
Moscow, Russia \\
aleksey.fedoseev@skoltech.ru}
\and
 
\IEEEauthorblockN{Miguel Altamirano Cabrera}
\IEEEauthorblockA{\textit{ISR Laboratory} \\
\textit{Skoltech}\\
Moscow, Russia \\
miguel.altamirano@skoltech.ru}
\and

\IEEEauthorblockN{Dzmitry Tsetserukou}
\IEEEauthorblockA{\textit{ISR Laboratory} \\
\textit{Skoltech}\\
Moscow, Russia \\
d.tsetserukou@skoltech.ru}

}

\maketitle

\begin{abstract}
The ability to perform locomotion in various terrains is critical for legged robots. However, the robot has to have a better understanding of the surface it is walking on to perform robust locomotion on different terrains. Animals and humans are able to recognize the surface with the help of the tactile sensation on their feet. Although, the foot tactile sensation for legged robots has not been much explored. This paper presents research on a novel quadruped robot DogTouch with tactile sensing feet (TSF). TSF allows the recognition of different surface textures utilizing a tactile sensor and a convolutional neural network (CNN). 
The experimental results show a sufficient validation accuracy of 74.37\% for our trained CNN-based model, with the highest recognition for line patterns of 90\%. In the future, we plan to improve the prediction model by presenting surface samples with the various depths of patterns and applying advanced Deep Learning and Shallow learning models for surface recognition.

Additionally, we propose a novel approach to navigation of quadruped and legged robots. We can arrange the tactile paving textured surface (similar that used for blind or visually impaired people). Thus, DogTouch will be capable of locomotion in unknown environment by just recognizing the specific tactile patterns which will indicate the straight path, left or right turn, pedestrian crossing, road, and etc. That will allow robust navigation regardless of lighting condition. Future quadruped robots equipped with visual and tactile perception system will be able to safely and intelligently navigate and interact in the unstructured indoor and outdoor environment.
\end{abstract}
\begin{IEEEkeywords}
Tactile Perception, CNN, Quadruped Robot
\end{IEEEkeywords}

\section{Introduction}
Tactile perception plays a crucial role in modern robotics, opening new frontiers in human-robot interaction and significantly increasing the environmental awareness of autonomous robots. In addition to visual estimation, humans and animals actively use tactile sensors in their skin and muscles to maintain balance and perform various agile motions \cite{b1, b2}. However, high attention has been brought to visual feedback systems in the field of legged robot locomotion, for instance, the laser range finder applied for surface adaptation by Plagemann et al. \cite{b24}, stereo-vision system proposed by Sabe et al. \cite{b3}, or infrared (IR) camera combined with ultrasound sensors proposed by Chen et al. \cite{b4}. 

\begin{figure}[ht]
 \includegraphics[scale=0.47]{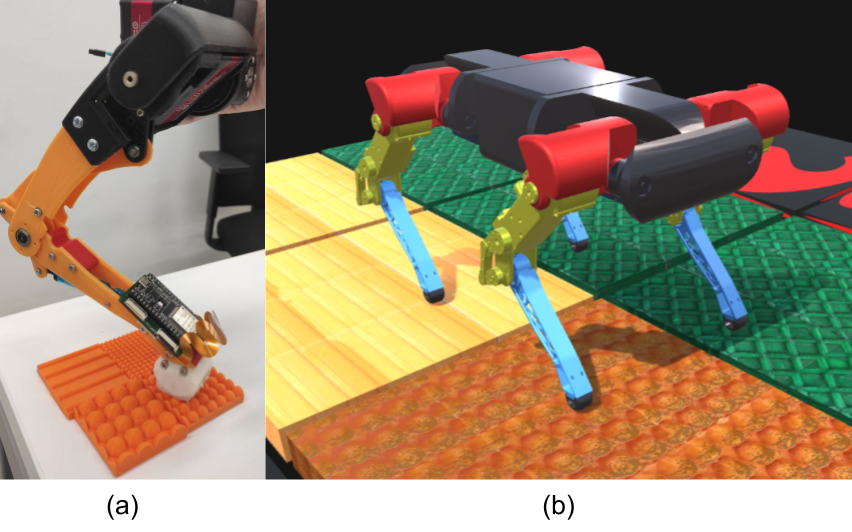}
 \centering
 \caption{(a) Robot dog contacting with the textured surface sample by tactile sensor array embedded in its foot. (b) VR concept scenario with recognized surface texture displayed to the remote operator.}
 \label{fig:1}
\end{figure}

Several works estimate the surface for legged robot locomotion through evaluation of joint position \cite{b5}. Camurri et al. \cite{b6} developed a Pronto state estimator for legged robots that can integrate pose corrections from the RGB camera, LIDAR, and odometry feedback. Sarkisov et al. \cite{b25} introduced a novel landing gear that allows surface profile estimation based on foot-pad IMU orientation and joint angles.
Zhang et al. \cite{b7} explored visual-based estimation of the tactile patterns by designing a robotic skin with a painted inner surface and installing a camera inside the robot leg. Smith et al. \cite{b8} suggested coupling data from foot contact sensors and Inertial Measurement Unit (IMU) to teach quadruped robot locomotion skills via Reinforcement Learning. 
A hybrid tactile sensor-based system was proposed by Luneckas et al. \cite{b9}, which are used in hexapod-legged robots to overcome obstacles. The sensor was designed to combine a limit switch and flexible polypropylene material that was connected with foot by silicone material, allowing the robot to estimate solid ground obstacles.
Legged robots are currently using direct feedback from the environment such as sonar, vision, LIDAR, and force feedback from joint actuators. Tactile sensors have recently been applied to expand the awareness of collaborative robots to its environment by feedback from a skin-like surface. In case of the legged robot, such sensors may be used beneath the robot's feet to estimate the properties of the surface. Adding tactile sensing to the robot’s feet can be beneficial for walking in challenging terrains in the same way haptic sensing plays an important role for animal locomotion in the nature. 


In this paper, we present the Touch Sensitive Foot (TSF), which is able to recognize the surface texture where the robot walks on with the help of trained CNN model. This research opens an efficient way to achieve environmental awareness for autonomous robots. So, the robot gait can be predetermined and the robot can walk on unknown terrains.
\newline

\section{Related Works}

The concept of haptic perception in robotic systems has been extensively applied in prototyping manipulators, mobile robots, underwater robots \cite{b10}, and drones. Several approaches towards surface estimation have been proposed, including integrated force and tactile sensors in the joints \cite{b11}, surface, and inner structures \cite{b12} of the robotic limbs. For example, Tsetserukou et al. \cite{b13} introduced a whole-sensitive robotic arm with optical torque sensors embedded in its joints. Contact force detection and control for robotic arm by joint torque sensors were investigated Dong et al. \cite{b14}. The proposed methods allow robots to efficiently estimate contact with surfaces, however, joint sensors are unable to measure fine details of the surface texture.

To estimate the distribution of forces during contact with the environment, a higher number of sensors should be embedded in robotic limbs. A pressure-sensitive skin that can be adapted to complex geometries was introduced by Fritzsche et al. \cite{b15} for safe human-robot interaction. This concept was further explored by Cheng et al. \cite{b16} presenting a humanoid robot with a sensor array on its surface. This system was enhanced with a low-resolution robot skin located on the bipedal robot soles suggested by Guadarrama-Olvera et al. \cite{b21} to reconstruct the supporting polygon and the pressure footprint online. The two-layer design of artificial robotic skin was suggested by Klimaszewski et al. \cite{b22}, which allows measuring the location, value, and direction of pressure from external force. Liu et al. \cite{b17} developed a large-scaled artificial sensitive skin for robots based on electrical impedance tomography. Dilibal et al. \cite{b18} proposed soft sensors for robotic grippers by a screen printing process with flexible material and ionic liquids. 

The aforementioned approaches allow to cover large areas with sensors, however, most of them lack the resolution necessary to be placed at the feet of a quadruped robot. There are, however, few displays that are developed to provide a high-resolution tactile data to the robot limbs. For example, a small thumb-sized vision-based sensor developed by Sun et al. \cite{b23} provides data with spatial resolution of 0.4 $mm$ and can be applied for dexterous manipulation. To obtain a high resolution of tilt estimation of the mobile charging robot, Okunevich et al. \cite{b19} proposed to evaluate tactile patterns. The developed vision-tactile perception system allows precise positioning of the charger end-effector and ensures a reliable connection between the electrodes of the two robots.

This paper presents a novel perception system for quadruped robots with CNN-based texture recognition from the data of a high-resolution tactile sensor array embedded in the sole of robotic foot. We evaluated system performance with eight 3D-printed samples of the surface. The proposed approach aims at improving the navigation of quadruped robots and their environmental awareness through special patterns placed on the surface and potentially teaching the adaptive locomotion of robots.

\section{DogTouch System Overview}
\begin{figure}[ht]
 \includegraphics[scale=0.42]{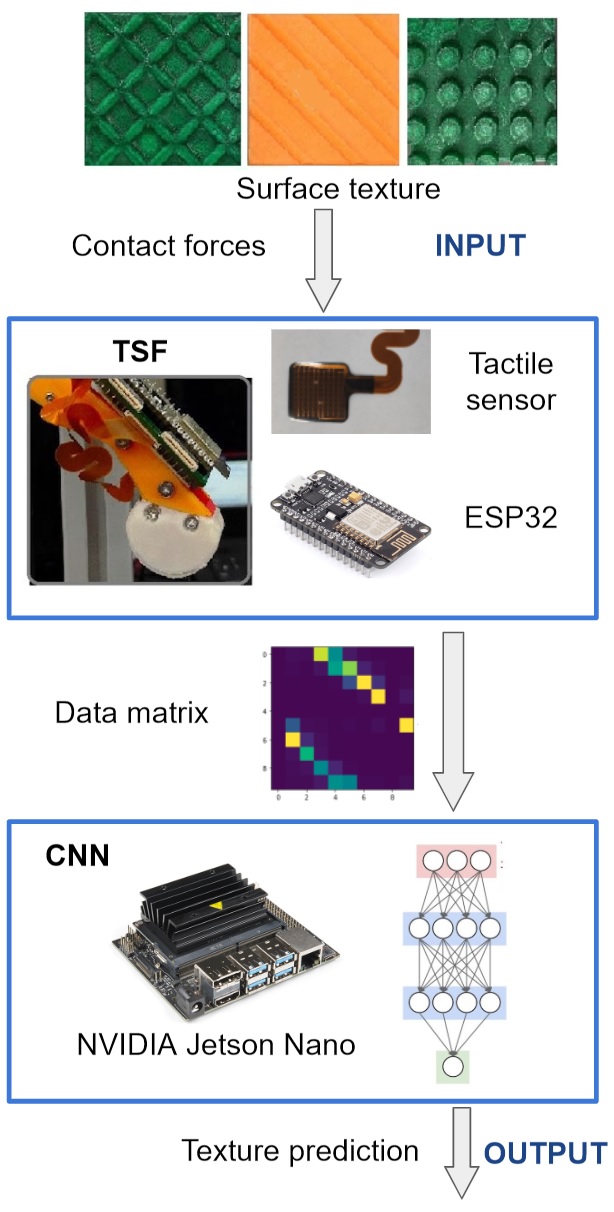}
 \centering
 \caption{Textures surface recognition with TSF. While contacting with the surface, the ESP32 microcontrollers process the data from the tactile sensor arrays. The data matrices are then transmitted to the NVIDIA Jetson Nano, which carries out the CNN-based texture recognition.}
 \label{fig:system architecture}
\end{figure}

All components of the system can be divided into three main modules, as shown in Fig. \ref{fig:system architecture}: Touch Sensitive Foot (TSF) with tactile sensor array, ESP32 microcontroller, and CNN model running on NVIDIA Jetson Nano to classify textures.  

The system works as follows: the ESP32 reads the tactile sensor arrays to recognize whether or not the sole touched the ground. When contact occurs, the ESP32 obtains the data matrix from the tactile sensor array and sends it to the CNN model running on Jetson Nano computer. The CNN model has been trained to recognize the surface textures. When the robot is aware of the texture type, the robot can localize itself (considering that the patterns are priory located in the specific configuration on the floor or pavement) and to optimize the gait to avoid slippage. 

\begin{algorithm}
\vspace{0.35cm}
\begin{flushleft}
\caption{Adaptive gait algorithm}
\While {robot is walking}
 { 
    read tactile sensor array data\;
    \If{foot touched the ground}
    {
    send tactile sensor array data to CNN\;
    estimate ground surface texture\;
        \eIf{Current Gait is suitable for predicted ground surface}
        {
        walk with current gait\;
        }
        {
        select the gait assigned to the surface pattern ID\; 
        walk with the selected gait\;
        }
    }
  }
\end{flushleft}
\end{algorithm}

\subsection{Leg Design of Quadruped Robot}
\begin{figure}[ht]
    \centering
    \includegraphics[scale=0.075]{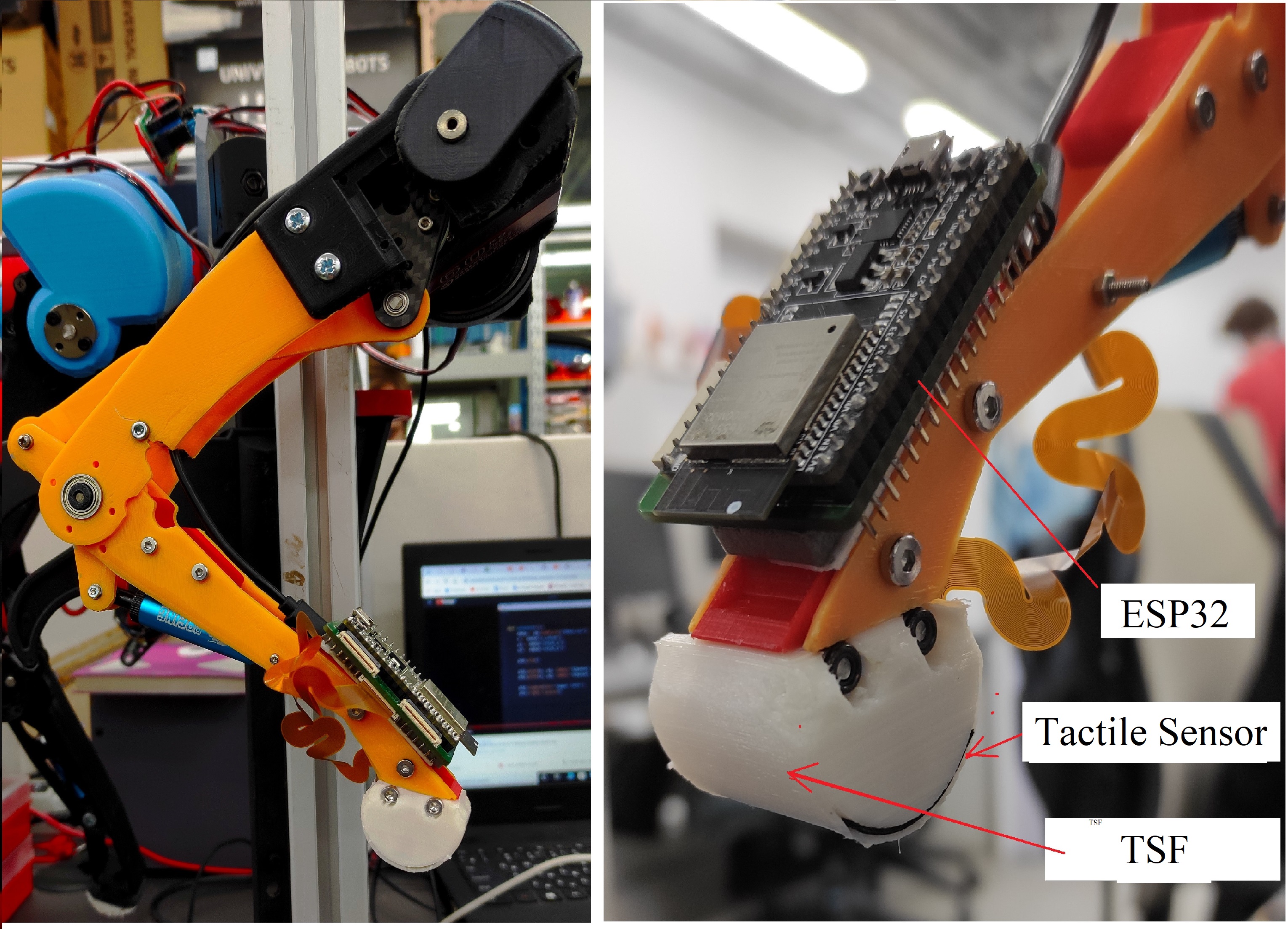}
    \caption{Touch Sensitive Foot (TSF) design with the embedded tactile sensor array.} \label{fig3}
\end{figure}

We have developed a unique customizable leg for a quadruped robot. 
The leg was designed to decrease the inertia, which is critical for the robot to have stable and efficient locomotion. 3D printed and carbon fiber parts were used for the fabrication of robotic legs. The manufactured legs have not only a lightweight structure but also a high strength. Each leg has 3 degrees of freedom: hip joint, upper leg joint and lower leg joint. Joints are actuated by RDS5160 SSG high torque digital servo motors with 7 $Nm$ maximum torque. Each servo motor is driven with 8.4 $V$ and 2.5 $A$ maximum current. The TSF was 3D printed using TPU (Thermoplastic polyurethane) material, which is flexible and strong enough to walk on harsh terrains. The tactile sensor (see \ref{subsection:sensors}) was installed in the sole as shown in Fig. \ref{fig3}.

 Once the foot touches the ground, the tactile sensor data is used to recognize surface texture with the CNN model.

\subsection{Embedded Tactile Sensor Array}
\label{subsection:sensors}

The TSF relies on the high-resolution tactile sensor array proposed by Yem et. al. \cite {b20}. The sensor is integrated into the soft sole of the robotic leg to provide the high-resolution perception of surface texture. High-resolution tactile sensor arrays allow the quadruped robot to collect detailed data of the textured surface. It is capable of sensing the maximum contact area of 5.8 $cm^2$ with a resolution of 100 points per frame. The sensing frequency is 120  $Hz$ (frames per second). The sensors allow the system to precisely detect the pressure on the small surface protrusions. The force detection range of the sensors is from 1 $N$ to 9 $N$.

\subsection{CNN Model for Tactile Perception}
\begin{figure}[ht]
 \includegraphics[width=\linewidth]{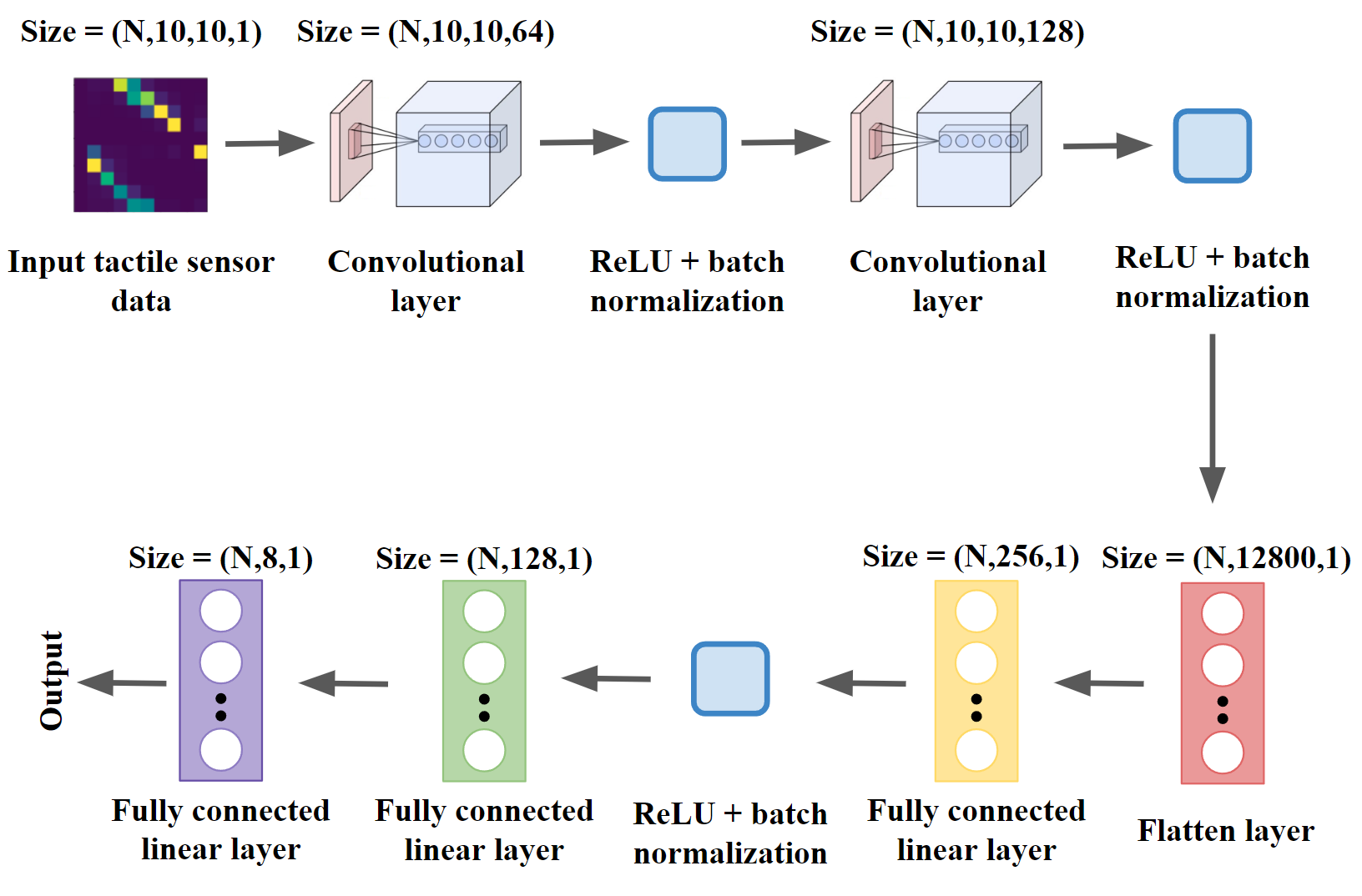}
 \centering
 \caption{CNN model for tactile perception system.}
 \label{fig:cnn model}
\end{figure}

CNN model consist of two convolutional layers with a 3x3 kernel and 3 fully connected linear layers with Rectified Linear Unit (ReLU) nonlinear activation functions, and batch normalization (see Fig. \ref{fig:cnn model}). Batch normalization acts as a regularizer and it speeds up the training of classification models. In addition, the batch normalization results in more predictive and well-behaved gradients being used in training, which eliminates major weight fluctuations and enables faster and more effective optimization.

The model receives the tactile sensor data as a three-dimensional matrix with the shape of $[1, 10, 10]$. The resolution of the data is relatively low in comparison with high-resolution camera frames or point cloud datasets. Therefore, our architecture does not include max pooling layers or strided convolutions in order to preserve the information. Such additions in the neural network architecture could be considered in the future with a higher number of tactile sensors or larger areas covered by tactile arrays. After convolutional layers, the data with the shape of $[batch size, 10, 10, 128]$ was flattened to one-dimensional vectors with 12800 elements. 


Finally, after three linear layers with output dimensions of 256, 128, and 8 (the number of texture types), the model output was received as the matrix with predictions for each class for all inputs in the batch.


\section{Texture Recognition Experiment}
\begin{figure}[ht]
\centering
 \includegraphics[width=0.95\linewidth]{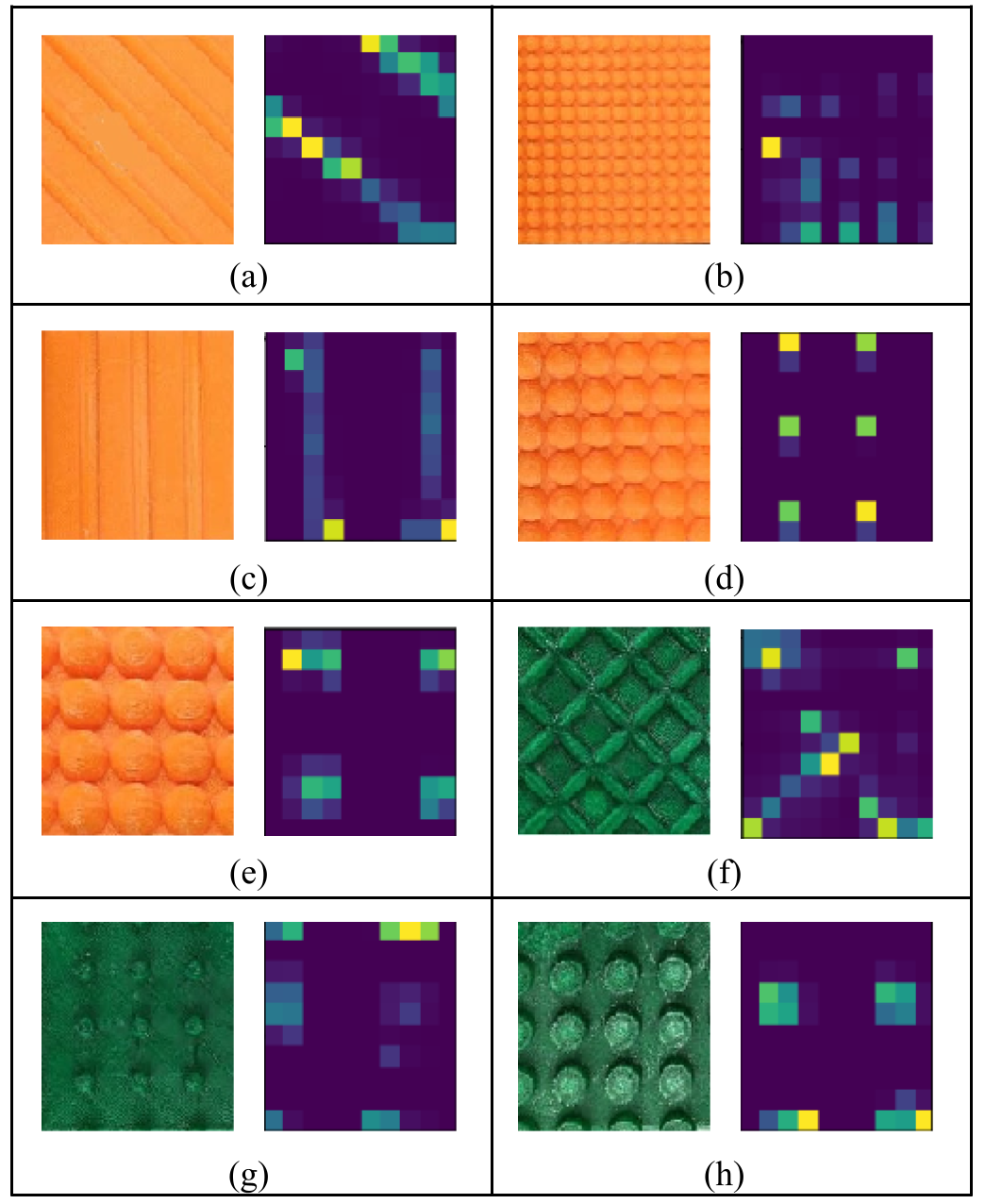}
 \centering
 \caption{3D-printed ground surface textures along with the corresponding detected tactile patterns by sensor array.}
 \label{fig4}
\end{figure}

The experiment was conducted with eight different textured patterns shown in Fig. \ref{fig4}. 

The following textured patterns were selected: diagonal lines with 1 $mm$ width and 5 $mm$ interval (Fig. \ref{fig4}a), dots of 1 $mm$ diameter with 1 $mm$ interval (Fig. \ref{fig4}b), vertical lines with 1 $mm$ width and 5 $mm$ interval (Fig. \ref{fig4}c), dots of 3 $mm$ diameter with 1 $mm$ interval (Fig. \ref{fig4}d), dots of 5 $mm$ diameter with 1 $mm$ interval (Fig. \ref{fig4}e), grid with 5 $mm$ interval (Fig. \ref{fig4}f), dots of 1 $mm$ diameter with 5 $mm$ interval (Fig. \ref{fig4}g), cylinders of 3 $mm$ diameter with 1 $mm$ interval (Fig. \ref{fig4}h). The sample of each ground surface pattern was 3D-printed with a size of 50x50 $mm^2$ from the PLA material.
The selected patterns vary in profile and resolution of the textures. In this research, we hypothesized that the proposed sensor arrays would allow us to obtain noticeable differences in the sensor readings with a minimal change of 2 $mm$ in the size of texture element.

\subsection{Dataset collection}
For training and validation of the CNN-based classification model, we collected a dataset including 800 data arrays from the tactile sensor (100 data arrays for 8 textured pattern). Dataset was divided into (90\%) for the train part, and (10\%) for the validation. TSF stepped on a certain textured plate 100 times, each time at a different angle. When there was a contact between the robotic leg and pattern, the system recorded the array in the dataset.


\subsection{Experimental Results}
\begin{figure}[ht]
 \includegraphics[width=0.8\linewidth]{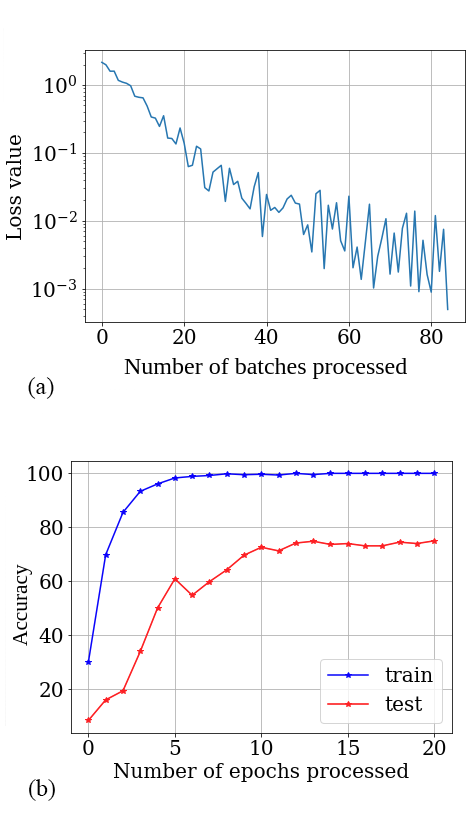}
 \centering
 \caption{(a) Training loss. (b) Training results and test validation of the developed CNN model. The accuracy does not increase after 12 epochs. }
 \label{fig:cnn results}
\end{figure}

The results of the CNN training are shown in Fig. \ref{fig:cnn results}. 

The training was conducted on a computer using the NVIDIA Tesla V100 GPU. Validation accuracy for our trained CNN-based model equals 74.37\%. After 12 epochs of training, accuracy does not change. Learning time was 12.6 $ms$ for the CNN-based model. 
After the experiment, we conclude that prediction for larger spheres is the same as for the cylindrical pattern. Line patterns demonstrated a high prediction rate of 90\%.

\section{Conclusions and Future Work}
A novel quadruped robot DogTouch is developed to leverage the tactile sensing of the robotic leg for the surface detection.  The proposed CNN-driven tactile perception system using data from tactile pressure sensors recognizes different textured patterns under the foot of the quadruped robot in 74.37\% cases in average.  The highest prediction rate of 90\% is achieved for line texture pattern.  
The neural network has sufficient accuracy in texture recognition. The sample, which is introduced into the network, is a part of a particular surface type with a certain texture. The cases, where the sample contains several types of texture, were not taken into account, but possible in the case of a continuous walk over an area (with a mixture of textures). With some modifications to the neural network presented in the paper (e.g. separating samples with texture mixes into additional classes or segmenting texture by its type and adding the dropout layers to reduce overfitting) higher accuracy of the texture prediction could be achieved in the future.


The proposed technology DogTouch can potentially considerably improve the robustness of navigation of legged robots regardless of the lighting conditions. Leveraging the sense of touch, such robots can navigate in unknown environment by reading the information from the tactile paving textured surface.  Additionally, robots will be capable to adapt the gait to the detected type of surface to avoid slippage.

\end{document}